\pdfoutput=1
\documentclass[11pt]{article}
\usepackage{subfig}
\usepackage{authblk}

\usepackage{ACL2023}

\usepackage{times}
\usepackage{latexsym}
\usepackage{amsmath}
\usepackage{svg}

\usepackage{graphicx}
\usepackage{subcaption}
\usepackage{makecell}
\usepackage{multirow}
\usepackage{algorithm}
\usepackage{algorithmic}
\usepackage{amsmath}
\usepackage{amsfonts}
\usepackage{booktabs}
\usepackage{siunitx}
\usepackage{float} 
\usepackage{enumitem} 
\usepackage{booktabs}
\usepackage{listings}
\usepackage{soul}
\usepackage{xcolor}
\usepackage{enumitem}

\usepackage{amsfonts}
\usepackage{bbm}
\usepackage[T1]{fontenc}
\usepackage[utf8]{inputenc}

\usepackage{microtype}

\usepackage{inconsolata}
\usepackage{tabularx}

\newcommand{\MethodName}{DuaShepherd}
\newcommand{\shepherd}{Math-Shepherd}

\title{DuaShepherd: Integrating Stepwise Correctness and Potential Rewards for Mathematical Reasoning}

\author[1]{\textbf{Yuanhao Wu}}
\author[1]{\textbf{Juntong Song}}
\author[2]{\textbf{Hanning Zhang}}
\author[2]{\textbf{Tong Zhang}}
\author[1]{\\\textbf{Cheng Niu}}
\affil[1]{NewsBreak}
\affil[2]{University of Illinois Urbana-Champaign}
\affil[ ]{ {\{yuanhao.wu,~cheng.niu\}@newsbreak.com}}

\begin{document}
\maketitle
\begin{abstract}
In this paper, we propose DuaShepherd, a novel reward modeling framework that integrates two complementary reward signals, correctness and potential, to enhance the mathematical reasoning capabilities of Large Language Models (LLMs). 
While correctness-based signals emphasize identification of stepwise errors, potential-based signals focus on the likelihood of reaching the correct final answer. 
We developed an automated pipeline for constructing large-scale reward modeling dataset with both signals. A unified, multi-head architecture was explored to train the two reward models in a multi-task setup, demonstrating benefits from learning both correctness and potential in parallel. By combining these two signals into a compound probability, our model achieves consistent performance improvements across multiple benchmarks.
Empirical evaluations on MATH500 and ProcessBench confirm that this combined reward significantly outperforms models trained on either reward type alone, achieving state-of-the-art performance under comparable resource constraints.
% Empirical evaluations on MATH500 and ProcessBench confirm that this combined reward framework significantly outperforms either reward type alone, and enable answer generators achieves state-of-the-art performance when comparing with other 7B LLMs.
% By unifying correctness-based and potential-based signals, we offer a robust and interpretable framework for reward modeling that achieves strong performance on challenging benchmarks.

\end{abstract}

\section{Introduction}
Large Language Models (LLMs) have achieved remarkable success across various NLP tasks \cite{DBLP:journals/corr/abs-2005-14165}, yet they continue to struggle in domains that require complex, multi-step reasoning, such as mathematical problem-solving \cite{hendrycksmath2021}. Recent advances have shown that prompting LLMs to think step-by-step, known as Chain-of-Thought (CoT) prompting, can significantly enhance their reasoning capabilities \cite{wei2023chainofthoughtpromptingelicitsreasoning}.

To further improve LLMs' internal capabilities in mathematical reasoning, reinforcement learning (RL) has been applied, yielding promising results. 
\citet{uesato2022solvingmathwordproblems} demonstrated that both outcome-based and process-based reward models achieve comparable performance on mathematical tasks. However, \citet{lightman2024lets} showed that with larger models and more fine-grained human-labeled data, process-based reward models (PRMs) can achieve significantly better results.

Despite its effectiveness, constructing step-wise supervision data, as described in~\citet{lightman2024lets}, requires substantial human effort to evaluate each step in an LLM-generated solution. Recent studies, like \shepherd, suggests that Monte Carlo sampling can be leveraged to estimate the probability of a given step leading to the correct final answer, offering a scalable alternative for building PRMs while maintaining competitive performance~\cite{wang-etal-2024-math}.

Besides its lower construction cost, we argue that synthetic PRM data generated via Monte Carlo sampling carries a different implicit meaning compared to OpenAI’s PRM800K data. 
Although PRM800K requires steps labeled as positive to progress towards the solution, when considering neutral and positive steps as a single class, their primary distinction from negative steps lies in the correctness of the step.
% While OpenAI's PRM data focuses on assessing the correctness of individual steps, 
% synthetic PRM data, such as that from \shepherd, emphasizes the likelihood or potential of a step leading to the correct final answer, effectively representing forward and backward reasoning directions. Based on this distinction, a key problem that arises is how to effectively integrate these two complementary yet distinct models into a unified scoring mechanism that leverages the strengths of both approaches.
On one hand, although PRM800K requires steps labeled as positive to progress towards the solution, when considering neutral and positive steps as a single class, their primary distinction from negative steps lies in the correctness of the step. On the other hand, PRM data obtained through Monte Carlo methods, such as \shepherd, emphasizes the likelihood or potential of a step leading to the correct final answer. Based on this distinction, a key challenge that arises is whether these two different signals can be effectively integrated into a more powerful process reward model that leverages the strengths of both approaches.

In this paper, we propose a novel approach, \MethodName, that effectively leverages these two different types of reward signals. Our experiments demonstrate that simply multiplying these two scores as a compound probability yields promising results. Besides, we develop an automated pipeline to construct a PRM training dataset that incorporates both types of reward signals.
% Using the multiplication score as supervision to distill the reward model leads to improved performance, achieving state-of-the-art results. This aligns well with OpenAI's weak-to-strong generalization paradigm\cite{burns2023weaktostronggeneralizationelicitingstrong}. 
We also find that a multi-task training schema, similar to ArmoRM \cite{wang2024interpretablepreferencesmultiobjectivereward}, where a single base model is shared to learn both PRM reward signals, ultimately enhances the reward models' overall performance. This further confirms that these two reward can complement each other.
% signals capture distinct aspects of the reasoning process and 
Our analysis shows that the two types of PRM reward signals indeed exhibit different reward patterns, and unifying them into a single framework leads to notable improvements in mathematical reasoning tasks.
By leveraging two existing datasets, PRM800K and \shepherd, and without introducing additional sampling or annotation, the DuaShepherd model proposed in this paper achieves significant improvements in MATH500 and ProcessBench, reaching a new state-of-the-art under comparable conditions.

These findings underscore the potential of combining multiple reward dimensions to build more capable reward models for reasoning.

Our main contributions are as follows: 
\begin{itemize}
    \item We propose a novel method that leverages two types of reasoning rewards: one that emphasizes identifying reasoning errors, and the other that focuses on evaluating the potential to arrive at the correct final answer..
    \item We present an automated dataset construction approach that enables large-scale generation of process supervision data with two types of reward signals for mathematical reasoning.
    \item We validate our reward fusion approach across multiple tasks and datasets, demonstrating its state-of-the-art performance.
\end{itemize}

\section{Related Work}

\subsection{LLM on Reasoning Task}
Despite demonstrating remarkable reasoning capabilities, particularly in solving math and coding problems \cite{huang-chang-2023-towards,NEURIPS2024_76ec4dc3,guan2025rstarmathsmallllmsmaster}, LLMs still remain suboptimal when dealing with complex reasoning tasks, e.g. overlooking logical fallacies or calculation errors \cite{zhong2024achieving97gsm8kdeeply, li2024reasonfallacyenhancinglarge, hendrycksmath2021}. Researchers have found that prompting LLMs to reason step by step, known as Chain-of-Thought prompting, significantly enhances their performance across various reasoning tasks, including mathematical problem-solving \cite{wei2023chainofthoughtpromptingelicitsreasoning, chu2024navigateenigmaticlabyrinthsurvey}. Similarly, self-reflection mechanisms, which allows LLMs evaluate and refine their own reasoning processes, have been demonstrated to enhance reasoning accuracy by identifying and correcting errors \cite{renze2024selfreflectionllmagentseffects}. Furthermore, integrating LLMs with external tools has been shown to further improve their mathematical reasoning capabilities \cite{gou2024toratoolintegratedreasoningagent, wang2023mathcoderseamlesscodeintegration}.

\subsection{Training Methods to Improve Mathematical Reasoning Abilities}
The mathematical reasoning capabilities of LLMs can be enhanced during two distinct training stages.

\paragraph{Pretraining} Pretraining LLMs on large amounts of math-related corpora has been proven effective in enhancing their mathematical reasoning abilities \cite{openai2024gpt4technicalreport, deepseekai2024deepseekv3technicalreport, grattafiori2024llama3herdmodels}.

\paragraph{Post-training} The pretrained model's capabilities can be further improved during the post-training stage. DeepSeekMath demonstrated that a carefully curated and diverse instruction-tuning dataset can significantly enhance mathematical reasoning, even through straightforward supervised fine-tuning \cite{deepseek-math}. Several studies have also shown that reinforcement learning is highly effective in strengthening LLMs' internal mathematical reasoning abilities \cite{uesato2022solvingmathwordproblems, lightman2024lets}.
% \paragraph{Inference-time scaling} 
Recent studies have indicated that through post-training, LLMs can exhibit emergent reasoning patterns, such as step-by-step reasoning and self-reflection, leading to significant performance improvements \cite{qwq-32b-preview, deepseekai2025deepseekr1incentivizingreasoningcapability}.

% One of the key advancements in reinforcement learning for improving LLMs' mathematical reasoning is the development of reward models, which guide optimization by providing structured feedback signals. This brings us to the distinction between outcome-based and process-based reward models, which play a crucial role in training models to reason effectively.
\begin{figure*}[!ht]
    \centering
    
    \includegraphics[width=0.95\textwidth]{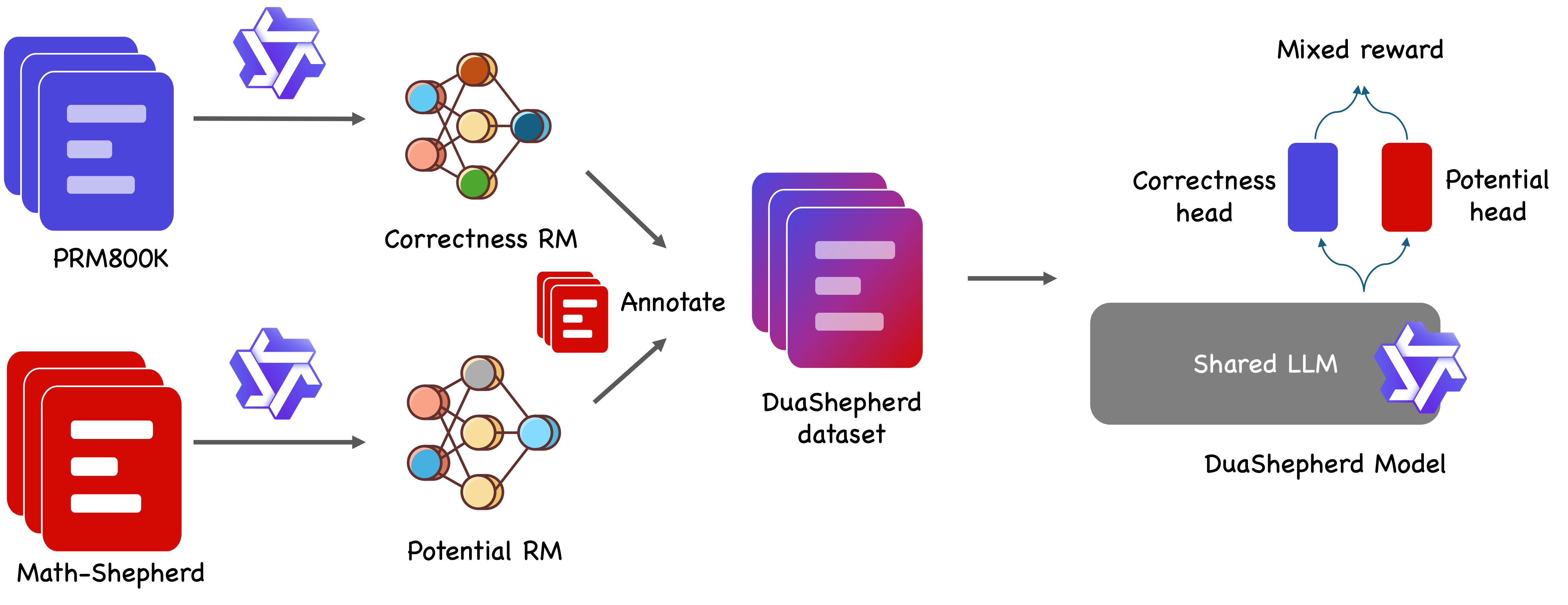}
    \caption{Overview of the \MethodName pipeline. We first trained reward models from PRM800K and \shepherd~dataset separately, then we used these reward models to automatically annotate the sampled reasoning trajectories to obtain the \MethodName~dataset with both potential and correctness reward labels. Finally, we trained a multi-head model with multi-task learning on the \MethodName~dataset. The \MethodName~approach yields substantial performance improvement.  }
    \label{fig:method}

\end{figure*}

\subsection{Reward Models}
% PRMs and ORMs
The reward model not only plays a crucial role in reinforcement training but also enhances the model's reasoning ability during inference stage by assisting in selecting high-quality generated results.
% DeepMind \cite{uesato2022solvingmathwordproblems} further categorizes reward models for language models into \textbf{outcome-based} and \textbf{process-based} reward models. 
In the domain of reasoning, reward models are typically categorized into two types: Outcome-based Reward Models (ORM) and Process-based Reward Models (PRM)~\cite{uesato2022solvingmathwordproblems}.
The ORMs provides reward signals based on the correctness of the final result and has been shown to significantly improve mathematical reasoning performance~\cite{cobbe2021trainingverifierssolvemath}. 
However, \citet{lightman2024lets} discovered that with larger models and more fine-grained human feedback signals, PRMs can achieve significantly better results. Nevertheless, collecting fine-grained human feedback often incurs substantial costs.

Fortunately, recent studies have demonstrated the feasibility of automatically constructing such datasets for PRM training. Math Shepherd \cite{wang-etal-2024-math, kazemnejad2024vineppounlockingrlpotential} introduced a novel approach that uses Monte Carlo sampling to automatically generate reward labels by evaluating whether a given trajectory leads to the correct result. Building on this idea, more sophisticated algorithms, such as Monte Carlo Tree Search (MCTS) \cite{chen2024steplevelvaluepreferenceoptimization}, have been proposed to generate step-wise preference data. Additionally, techniques like self-reflection have been incorporated into tree search methods to dynamically enhance the quality of responses \cite{zhang2024accessinggpt4levelmathematical}. \citet{setlur2024rewardingprogressscalingautomated} took this concept a step further by not only assessing whether the current step can lead to the final result but also evaluating whether each additional step meaningfully contributes to the overall success rate.
% Carnegie Mellon University
Moreover, recent work from \citet{setlur2024rlincorrectsyntheticdata} has explored an alternative approach by directly constructing both correct and incorrect trajectories in mathematical reasoning. By leveraging reinforcement learning on these structured datasets, they demonstrated that training on per-step incorrect responses can significantly improve model generalization.

\section{Method}

\subsection{PRM} PRMs enhance LLMs by evaluating each step of the reasoning process rather than just the final outcome. This stepwise supervision enables the identification and correction of errors as they occur, leading to improved performance on complex tasks.

The PRM is often trained with the following loss function:

\begin{equation}
    \mathcal{L} = - \sum_{i=1}^K[y_{s_i} \log r_{s_i} + (1 - y_{s_i}) \log (1 - r_{s_i})]
\end{equation}
% _{\text{PRM}}
\noindent where \( y_{s_i} \in [0,1] \) represents the correct label (golden answer) for \( s_i \), the \( i \)-th reasoning step of solution \( s \). The value \( r_{s_i} \) denotes the sigmoid score assigned to \( s_i \) by PRM, and \( K \) is the total number of reasoning steps involved for \( s \)~\cite{wang-etal-2024-math}.
In this study, we also utilize this loss function to train the models.

\subsection{Construction of the \MethodName~Dataset}
\label{sec:PRT}
As shown in Figure~\ref{fig:method}, to construct a dataset with both potential and correctness reward labels, we leverage two existing datasets, PRM800K~\cite{lightman2024lets} and \shepherd~\cite{wang-etal-2024-math}, along with reward models trained on each dataset separately.

First, we trained a correctness reward model on the PRM800K dataset. According to original label definition, during training, we treated both the neutral label and the positive label as \(y=1\)~(correct), while the negative label was considered \(y=0\)~(incorrect). Next, we used the \shepherd~dataset to train the potential reward model with its original binary label. 

Unlike the PRM800K dataset, which requires substantial manual annotation costs, the \shepherd~dataset has a significant advantage in that it allows for the automated generation of large volumes of high-quality data. To preserve this efficiency in our approach, we employed the model trained with the PRM800K dataset to predict the correctness of each reasoning step in the \shepherd~dataset. This process generates pseudo-labels for the correctness reward. To simultaneously learn both reward using a single dataset, soft labels are preferred to prevent prematurely truncating potential trajectories during joint training. Since the \shepherd~dataset only provides binary hard labels, we used the potential reward model to generate corresponding pseudo-soft labels as potential rewards.

Through this approach, we obtained independent correctness and potential reward labels for every sample in the \shepherd~dataset. We named this dataset \MethodName, and used this dataset to train reward models combining both correctness and potential. 
Specifically, since the \MethodName~dataset is constructed using Monte Carlo sampling and pseudo-labeling, its scale is easily expandable.

% \MethodName~II: 
\subsection{The \MethodName~Model}
We employ a multi-head model in conjunction with multi-task learning to simultaneously learn two types of reward labels, as shown in Figure~\ref{fig:method}. Specifically, we add separate fully connected layers upon a shared base LLM to independently predict two distinct rewards:

\begin{equation}
    \label{rc}
    R_{\text{correctness}} = \sigma(\mathbf{w}^T_{1}\mathbf{h}+\mathbf{b}_1)
\end{equation}
\begin{equation}
    \label{rp}
    R_{\text{potential}} = \sigma(\mathbf{w}^T_{2}\mathbf{h}+\mathbf{b}_2)
\end{equation}

\noindent where \( \sigma \) denotes the sigmoid function, while \( \mathbf{w}_1, \mathbf{b}_1 \) and \( \mathbf{w}_2, \mathbf{b}_2 \) represent the weights and biases of two fully connected layers, respectively. The variable \( \mathbf{h} \) refers to the hidden state obtained from the last layer of base LLM.

This design leverages the shared foundational LLM to capture general language representations, while the individual fully connected layers specialize in estimating each specific reward. For the multi-head model, we use the sum of the binary cross-entropy losses from the two rewards as the training loss:
\begin{equation}
    \mathcal{L}_\text{mixed} = \mathcal{L}(y_\text{c.}, r_\text{c.}) + \mathcal{L}(y_\text{p.}, r_\text{p.}) 
\end{equation}

% \MethodName~I: 
% \subsection{Compound Probability as Mixed Reward}

Multiple dimensions of rewards offer rich and detailed signals. \( R_{\text{correctness}} \) evaluates past reasoning steps, representing the probability that the existing steps are correct. Meanwhile, \( R_{\text{potential}} \) looks ahead, estimating the likelihood that the reasoning will lead to a successful final solution. 
Given the orthogonal nature of these two metrics, we propose to use the compound probability as the final \MethodName~reward applying the chain rule of probability.
% Given the orthogonal nature of these two metrics, the reward model, \MethodName~I, is proposed as a compound probability, applying the chain rule of probability.

%Although multiple dimensions of rewards can provide rich and detailed signals, in practical applications, we aim to obtain a reward model that delivers a single reward signal. Therefore, it is necessary to combine the two rewards.

%We believe that incorrect steps often indicate low-quality reasoning chains. Therefore, we aim for the model to improve the probability of arriving at the correct final answer while maintaining accuracy throughout the reasoning process. 
%Under the scenario of manually annotated binary correctness rewards , the mixing method can be expressed as follows:  

%\begin{equation}
%R_{\text{mixed}} = \mathbbm{1}_{R_{\text{correctness}}\ne-1} \cdot R_{\text{potential}}
%\end{equation}

%\noindent here, \( R_{\text{correctness}} \) is assigned a value of -1 to indicate incorrect reasoning steps, \( R_{\text{potential}} \) represents the potential to achieve a correct final answer. In general, if we denote \( R_{\text{correctness}} \) as the probability of obtaining correct reasoning steps, then \( R_{\text{mixed}} \) can be expressed as the product of two rewards:

\begin{equation}
R_{\text{\MethodName}} = R_{\text{correctness}} \cdot R_{\text{potential}}
\end{equation}

\begin{table*}[ht!]
\small
\centering
\begin{tabular}{cllc}
\toprule
\multirow{2}{*}{\textsc{Generator}}           & \multirow{2}{*}{\textsc{No.}} & \multirow{2}{*}{\textsc{Model}}               & \multirow{2}{*}{\textsc{Accuracy}} \\
                                      &                      &                                      &                           \\
\midrule
\multirow{6}{*}{\makecell{\textbf{Mistral-7B: MetaMath}\\pass@1: 0.266\\maj@64: 0.362\\pass@64: 0.750}} & 1                    & Qwen2.5-Math-7B: PRM800K~(our trained)    &          0.472                 \\
                                      & 2                    & Qwen2.5-Math-7B: Math-Shepherd~(our trained)           &          0.472                 \\
                                      & 3                    & Mixing 1 \& 2 with compound probability                  &           0.498                \\
                                      & 4                    & \MethodName~(ours) &      \textbf{0.526}                     \\
                                      &                     & \MethodName~correctness head~(ours) &      0.506                     \\
                                      &                     & \MethodName~potential head~(ours) &      0.504                     \\

                                      % & 5                    & Llama3.1-8B-PRM-Deepseek-Data~\cite{xiong2024rlhflowmath} &      0.45                     \\

\midrule
\multirow{6}{*}{\makecell{\textbf{DeepSeekMath-Instruct-7B}\\pass@1: 0.412\\maj@64: 0.556\\pass@64: 0.862}}     & 1                    & Qwen2.5-Math-7B: PRM800K~(our trained)    &        0.594                   \\
                                      & 2                    & Qwen2.5-Math-7B: Math-Shepherd~(our trained)           &      0.544                     \\
                                      & 3                    & Mixing 1 \& 2  with compound probability                 &        0.602                   \\
                                      & 4                    & \MethodName~(ours) &        0.622                 \\
                                      &                     & \MethodName~correctness head~(ours) &        \textbf{0.626}                 \\
                                      &                     & \MethodName~potential head~(ours) &        0.560                 \\
\midrule
\multirow{6}{*}{\makecell{\textbf{Qwen-2.5-Math-Instruct-7B}\\pass@1: 0.540\\maj@64: 0.840\\pass@64: 0.890}}     & 1                    & Qwen2.5-Math-7B: PRM800K~(our trained)    &        0.808                   \\
                                      & 2                    & Qwen2.5-Math-7B: Math-Shepherd~(our trained)           &      0.820                     \\
                                      & 3                    & Mixing 1 \& 2  with compound probability                 &        \textbf{0.824}                   \\
                                      & 4                    & \MethodName~(ours) &        \textbf{0.824}                  \\
                                      &                     & \MethodName~correctness head~(ours) &        0.804                 \\
                                      &                     & \MethodName~potential head~(ours) &        \textbf{0.824}                \\

\bottomrule
\end{tabular}
\caption{Best-of-N accuracy of different generators on MATH500 with different verification models. For each prompt, we sampled 64 solution candidates. We also provide the pass@1, maj@64~(majority voting among 64 samplings) and pass@64~(upper bound) accuracies of each generator.
% The upper bound accuracy~(pass@64) of Mistral-7B, DeepSeekMath-Instruct-7B and Qwen-2.5-Math-Instruct-7B are 75.0\%, 86.2\% and 82.4\% respectively.
}
    \label{tab:bestofn}
\end{table*}

\section{Experimental Setup}
\subsection{Evaluation Datasets and Metrics} We conduct our experiments using two distinct datasets. The first dataset is the widely adopted MATH benchmark \citep{hendrycksmath2021}, which serves as a standard for evaluating mathematical problem-solving capabilities. The second dataset, ProcessBench~\cite{zheng2024processbenchidentifyingprocesserrors}, is specifically designed to assess mathematical reasoning within the PRM framework.

For the MATH dataset, in the verification scenario, we leverage a computationally efficient subset, MATH500, which corresponds precisely to the test set used in \citet{lightman2024lets} and \citet{wang-etal-2024-math}. 
To evaluate different reward models, a generator produces 64 candidate solutions for each test problem. Each reward model then selects the highest-scoring solution from the generated candidates. The final evaluation reports the average best-of-N accuracy across different test problems for each method.

In ProcessBench, models are required to detect the earliest step that contains an error or confirm that all steps are correct. We generally follow the evaluation instructions provided by the authors. However, for models that directly output a scalar value, we do not select the threshold based on the GSM8K subset of the data. Instead, we fix it at 0.5.
% , which aligns with standard binary classification
% We make this choice because we found that threshold selection based on grid search is highly unstable, with performance fluctuating significantly depending on the granularity of the search. 
We then report the overall performance based on this standardized procedure. 
 % to ensure consistency with its established evaluation protocol

Therefore, our evaluations assess both the reward model's ability to select the correct final solution and its capability to identify errors in the reasoning process.

\subsection{Parameter Setting}
To construct a dataset with both potential and correctness reward labels, the \( R_{\text{correctness}} \) model and the \( R_{\text{potential}} \) model were fine-tuned on the PRM800K and \shepherd~datasets respectively for a total of one epoch based on the Qwen-2.5-Math-7B model~\cite{qwen2.5} using a learning rate of \(2 \times 10^{-5} \). 

The final \MethodName~reward model was also trained from the Qwen-2.5-Math-7B model with a learning rate of \(2 \times 10^{-5}\).

% and the 
Three 7B models, Mistral-7B~\cite{jiang2023mistral7b}, DeepSeekMath-Instruct-7B~\cite{deepseek-math} and Qwen-2.5-Math-Instruct-7B~\cite{yang2024qwen2} are used as answer generators in the experiment. Following the practice in \shepherd, we fine-tune Mistral-7B on MetaMATH~\cite{yu-etal-2024-ovm} for three epochs to further improve its mathematical reasoning capability. A learning rate of \(5 \times 10^{-6}\) is used in the training. DeepSeekMath-Instruct-7B and Qwen-2.5-Math-Instruct-7B are specifically designed for mathematical reasoning, and so no further fine-tuning is needed. We selected these three models to evaluate our reward model's effectiveness across generators with varying reasoning abilities.

We conduct our experiments using 8 NVIDIA H100 GPUs with a total of 8*80 GB of GPU memory.

\section{Experimental Results}

% \usepackage{multirow}
% \begin{table*}[ht!]
% \small
% \centering
% \begin{tabular}{lllllcl}
% \toprule
% \multirow{3}{*}{\textsc{No.}} & \multicolumn{4}{c}{\textsc{Training Data}}                               & \multirow{3}{*}{\textsc{Model}} & \multirow{3}{*}{\textsc{ProcessBench}} \\
%                      & \multicolumn{2}{c}{PRM800K} & \multicolumn{2}{c}{Math-Shepherd} &                        &                               \\
%                      & C. Reward    & P. Reward    & C. Reward       & P. Reward       &                        &                               \\ 
% \midrule

% 1                    & Original     & -            & -               & -               & Qwen2.5-Math-7B        &                               \\
% 2                    & -            & -            & -               & Original        & Qwen2.5-Math-7B        &                               \\
% 3                    & Original     & -            & -               & Original        & Multiply of 1 \& 2     &                               \\
% 4                    & -            & -            & Distilled       & Original        & Qwen2.5-Math-7B        &                               \\
% 5                    & -            & -            & Distilled       & Distilled       & Qwen2.5-Math-7B        &                              \\
% \bottomrule
% \end{tabular}
% \caption{}
% \label{table:processbench}
% \end{table*}

% \usepackage{multirow}
\begin{table*}[ht!]
\small
\centering
\begin{tabular}{lccccc}
\toprule
 {\textsc{Model}}               & \textsc{GSM8K} & \textsc{MATH} & \textsc{\makecell{Olympiad-\\Bench}} & \textsc{\makecell{Omni-\\MATH}} &{\textsc{Avg.}} \\
\midrule
\multicolumn{6}{c}{Process Reward Models}\\
\midrule
Skywork-PRM-7B$^*$~\cite{skyworkopeno12024}      & 70.8 & 53.6 & 22.9 & 21.0                & 42.1                          \\
Llama3.1-8B-PRM-Deepseek-Data$^*$~\cite{xiong2024rlhflowmath} &  38.8 & 33.8 & 16.9 & 16.9 & 25.6 \\
Llama3.1-8B-PRM-Mistral-Data$^*$~\cite{xiong2024rlhflowmath} &  50.4 & 33.4 & 13.8 & 15.8 & 28.4 \\
% Llama3.1-8B: PRM800K~(our trained) & 20.7 & 36.1 & 29.7 & 29.3 & 29.0 \\
% Llama3.1-8B: Math-Shepherd~(our trained) & 28.5 & 12.1 & 4.3 & 4.2 & 12.3 \\
% Llama3.1-8B: \MethodName~(ours) & 63.9 & 45.8 & 34.6 & 39.3 & 45.9 \\
% Qwen2.5-Math-1.5B: PRM800K~(our trained) & 47.6 & 65.0 & 55.2 & 57.4 & 56.3 \\
% Qwen2.5-Math-1.5B: Math-Shepherd~(our trained) & 55.4 & 26.0 & 10.4 & 8.6 & 25.1 \\
% Qwen2.5-Math-1.5B: \MethodName~(ours) & 68.3 & 66.7 & 57.8 & 58.4 & 62.8 \\

Qwen2.5-Math-7B: PRM800K~(our trained)    & 61.2 & 66.1 & 51.6 & 51.0 &       57.5                      \\
Qwen2.5-Math-7B: Math-Shepherd~(our trained)    &  60.3 & 27.8 & 8.9 & 4.1 &         25.3                   \\
Mixing above two models' rewards with compound probability      & 74.5 & 67.7& 54.7& \textbf{55.2}   &            63.0                   \\
Qwen2.5-Math-7B: \MethodName~(ours) & \textbf{78.7}  & \textbf{68.3} & \textbf{60.2} & 54.9 &           \textbf{65.5}                 \\
Qwen2.5-Math-7B: \MethodName~correctness head~(ours) & 63.9  & 64.2 & 55.2 & 52.7 &          59.0                 \\
Qwen2.5-Math-7B: \MethodName~potential head~(ours) & 65.0  & 29.5 & 9.5 & 4.9 &           27.2                \\

\midrule
\multicolumn{6}{c}{Language models, prompted as Critic Models}\\
\midrule
Qwen2.5-72B-Instruct$^*$~\cite{qwen2.5}     & 76.2 & 61.8 & 54.6 & 52.2           & 61.2                          \\
Qwen2.5-Math-72B-Instruct$^*$~\cite{qwen2.5}  & 65.8 & 52.1 & 32.5 & 31.7 & 45.5                          \\
QwQ-32B-Preview$^*$ ~\cite{qwq-32b-preview}  & 88.0 & 78.7 &  57.8 & 61.3 & 71.5                          \\
GPT-4o-0806$^*$     & 79.2 & 63.6 & 51.4 & 53.5                    & 61.9                          \\
o1-mini$^*$         & \textbf{93.2} & \textbf{88.9} & \textbf{87.2} & \textbf{82.4}                & \textbf{87.9}                          \\
\bottomrule
\end{tabular}
\caption{Performance of different models on ProcessBench. Our model significantly outperforms PRM models of the same size and even surpasses many large critic models with long reasoning chains. * indicates results sourced from the ProcessBench paper. }
\label{table:processbench}
\end{table*}
\subsection{Effectiveness of Compound Probability}
\label{sec:rm}
As shown in Table~\ref{tab:bestofn}, by comparing experiments 1, 2, and 3 for all the generators, it can be observed that compounding the rewards from two separately trained reward models yields better results than either reward model alone. For the \MethodName~model, this trend holds for two out of the three generators, achieving performance comparable to or better than the stronger head. The exception is DeepSeekMath, where the significant performance gap between the two heads results in the compounded reward performing worse than the best individual head. 

As shown in Table~\ref{table:processbench}, in the ProcessBench dataset, the model trained with PRM800K demonstrates a significant advantage over the model trained with Math-Shepherd. By multiplying the two rewards, the overall performance improves from 57.5\% of the PRM800K model to 63.0\%. The compound probability of the \MethodName~model also shows significant improvement over each head of the model. 

\subsection{Effectiveness of Multi-Task Training}
\label{sec:mtl}
We can observe that multi-task training further enhances the capability of the reward model. Although the two reward signals focus on different aspects of reasoning steps, they seem to positively influence each other.

In the MATH500 dataset with the Mistral-7B generator, using only the correctness head of Model 4 as the verifier, we achieve an accuracy of 0.506, which is an improvement of 3.4\% over Model 1.
Using only the potential head as the verifier, the accuracy reaches 0.504, surpassing its counterpart, Model 2, by 3.2\%.
By comparing Experiments 3 and 4, when the rewards from both heads are multiplied to form the final verification score, the accuracy reaches 0.526, which improves upon the multiplication of labels used to train this model (i.e., Experiment 3) by 1.8\%. 
These results also hold for more powerful generators, namely DeepSeekMath-Instruct-7B and Qwen-2.5-Math-Instruct-7B. 
% The \MethodName~model also shows advantage over multiplying rewards from two separate models in the ProcessBench dataset.

For the ProcessBench dataset, the correctness head of \MethodName~achieves an average F1 score that surpasses its teacher model by 1.5\%. Additionally, the potential head outperforms its teacher model with an average F1 improvement of 1.9\%.

These results highlight the benefits of joint optimization over the two reward signals, demonstrating that multi-task learning enables the reward model to capture richer reasoning patterns and enhance verification accuracy.

% In the MATH 500 dataset, when using Mistral-7B as the generator, multiplying the two rewards leads to a 2.6\% improvement in accuracy. Similarly, when sampling results with a more powerful generator, DeepSeekMath-Instruct-7B, we observe a comparable accuracy gain. Specifically, the accuracy obtained by multiplying the two rewards surpasses that of Model 1 and model 2 by 2.6\% and 5.8\%, respectively.

\subsection{Performance with Different Generators}
% As shown in Table~\ref{tab:bestofn}, the conclusions regarding MATH500 in Section~\ref{sec:rm} and Section~\ref{sec:mtl} exhibit strong consistency across both the weaker generator, Mistral-7B, and the stronger generator, DeepSeekMath-Instruct-7B. 
% The corresponding verification approach demonstrates a consistent accuracy gap of approximately 10\% on the two generators. This result indicates that our method remains effective even for models with strong reasoning capabilities, leading to a significant improvement in reasoning accuracy.
As shown in Table~\ref{tab:bestofn}, for all generators, \MethodName~achieves an accuracy significantly higher than pass@1. This result indicates that our method remains effective even for models with strong reasoning capabilities, leading to a substantial improvement in reasoning accuracy. However, it is also important to note that as the generator’s inherent capability improves, the advantage of \MethodName!over majority voting diminishes. Although \MethodName~consistently achieves higher accuracy than models trained on the two baseline datasets across all generators, we observe that as the generator’s capability improves, the performance gain of \MethodName~ also narrows. 

We argue that this phenomenon arises because the solutions in the training data for the PRM model were generated by relatively outdated models, leading to a significant discrepancy between these solutions and those produced by the latest generators. Specifically, we observe that solutions generated by Qwen-2.5-Math-Instruct-7B contain a substantial amount of Chinese text and garbled characters, which are absent in the PRM800K and \shepherd~datasets. Using a model more aligned with the target generator to sample and generate PRM training data should improve the performance of the PRM model.

\subsection{Performance with Different Number of Candidate Solutions}
\begin{figure}[t]
\includegraphics[width=0.5\textwidth]{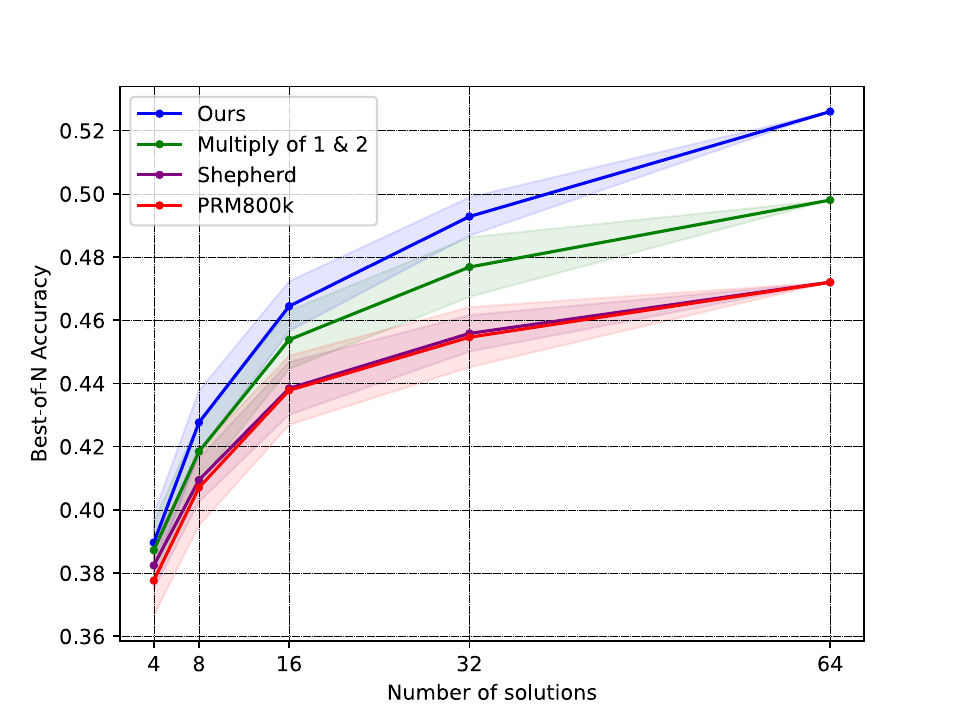}
\caption{Performance of Mistral-7B using different verification models across different numbers of solution candidates on MATH500.}
\label{fig:bestofn}
\end{figure}
\begin{figure*}[!ht]
    \centering
    \subfloat[Math-Shepherd]{\includegraphics[width=0.33\linewidth]{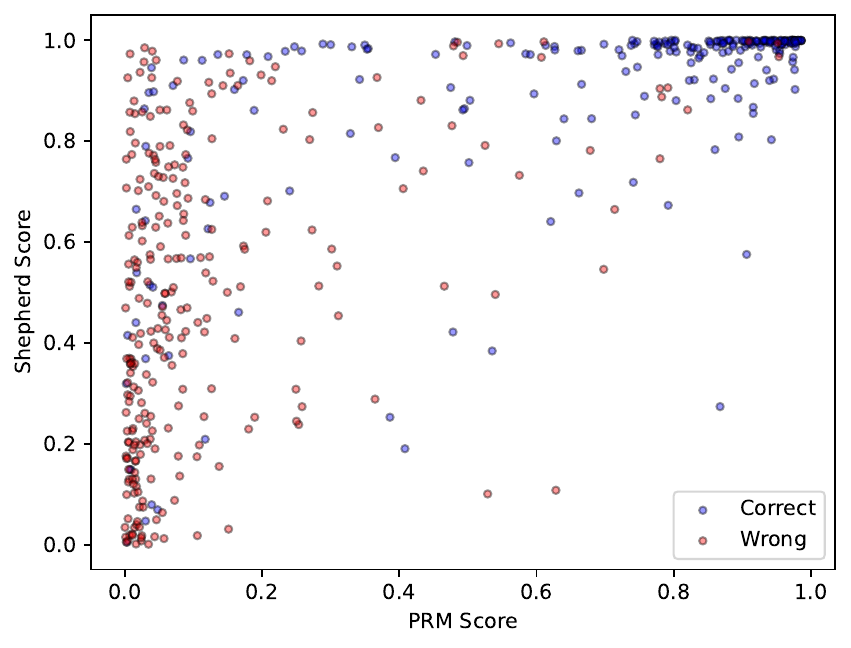}}
    \subfloat[PRM800K]{\includegraphics[width=0.33\linewidth]{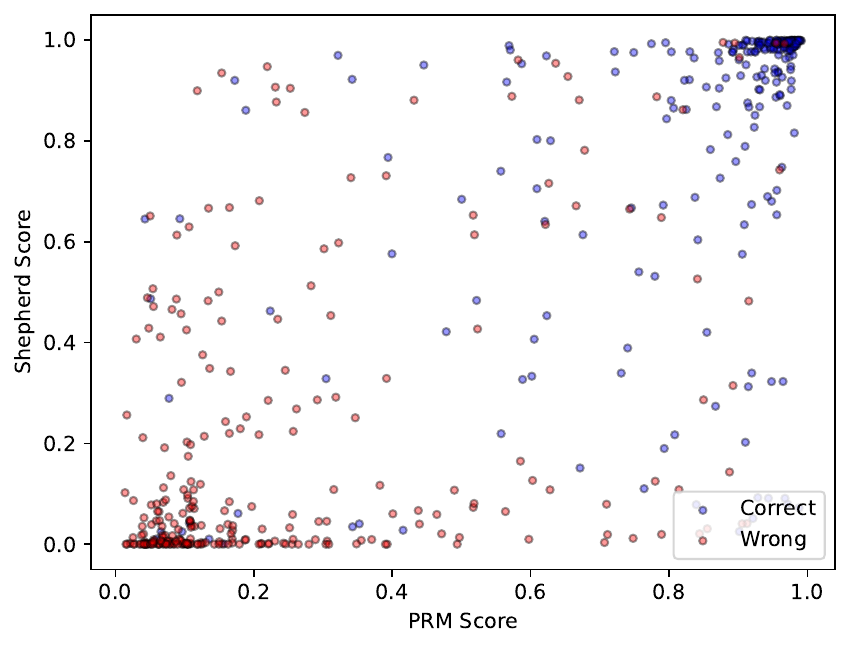}}
    \subfloat[\MethodName]{\includegraphics[width=0.33\linewidth]{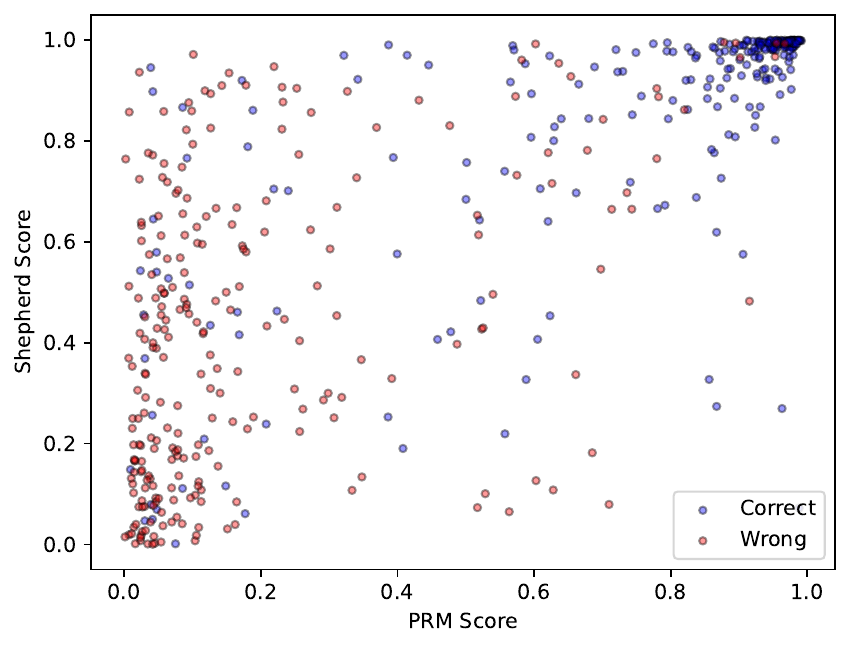}}

    \caption{Reward distribution of the best solutions picked by different reward models in the MATH500 test set. As a verifier, Math-Shepherd exhibits most of its errors near the \( P_{\text{correctness}} = 0 \) region. As for the PRM800K model, it focuses on correctness; however, many of its errors arise from failing to select solutions with high \( P_{\text{potential}} \) when correctness probabilities are generally low. By integrating these two reward signals, \MethodName~effectively mitigates both types of errors, leading to a significant improvement in performance.}
    \label{fig:three-max}

\end{figure*}

Figure~\ref{fig:bestofn} presents a performance comparison of different strategies applied to varying candidate solution counts, ranging from 1 to 64, sampled from Mistral-7B generator on the Math500 benchmark. When varying the number of candidate solutions, combining the rewards from PRM800K and Math-Shepherd yields higher accuracy than using either reward individually, and our \MethodName~model consistently further enhances this accuracy.

\section{Discussion}

\subsection{Quality of the DuaShepherd Dataset}

As shown in~Table~\ref{tab:quality}, We conducted a quantitative evaluation of our annotating models, and the results suggest that they are of high quality:

\paragraph{Potential RM} For pseudo-labels generated by the potential reward model, we computed Precision, Recall, and F1 score against the original MathShepherd binary labels. Although these labels were themselves generated via Monte Carlo sampling, they provide a strong baseline. The pseudo-labels from our model achieved an F1 score of 0.9346, demonstrating high consistency with the original labeling.

\paragraph{Correctness RM} For pseudo-labels from the correctness reward model, we constructed reference annotations using OpenAI's o1-mini model on 1,000 randomly sampled reasoning trajectories, following the prompt from the ProcessBench paper. This model has been reported to achieve an F1 score of 87.9 in identifying the first error in a reasoning trajectory, making it a reliable proxy for human judgment. Our pseudo-labels—defined as the steps leading up to the first error—achieved an F1 score of 0.796 against the o1-mini annotations. This result further supports the effectiveness of our correctness reward model in capturing critical reasoning signals.

\begin{table*}[t]
\small
\centering
\begin{tabular}{lllll}
\toprule
\textsc{Model}      & \textsc{Ground Truth} & \textsc{Precision} & \textsc{Recall} & \textsc{F1} \\
\midrule
Potential RM        &   MathShepherd      &    94.16\% & 92.77\% & 93.46\%        \\
Correctness RM  &   o1-mini annotation      &    80.00\% & 79.20\% & 79.60\%         \\

\bottomrule
\end{tabular}
\caption{Performance of annotating models.}
    \label{tab:quality}
\end{table*}

These metrics indicate that our pseudo-labels are not only aligned with strong automated baselines but also capture reasoning correctness with a high degree of reliability. This high label quality is crucial to the effectiveness of our reward modeling and downstream applications.

\subsection{Relationship of the two Rewards}
We calculated the Pearson correlation coefficient between the two rewards of Mistral-7B MATH500 solutions candidates. The correlation coefficient for correct solutions and incorrect solutions are 0.712 and 0.392 respectively. The overall correlation coefficient is 0.812.

We plotted the best Mistral-7B solutions by different reward models of MATH500 test set in Figure~\ref{fig:three-max}. 
Among all the reward models, points representing incorrect solutions cluster in the lower-left quadrant, while those representing correct solutions concentrate in the upper-right quadrant. This distribution indicates a general consensus between the two reward metrics in distinguishing correct from incorrect solutions. 
In Figure~\ref{fig:three-max}(a), when only the Math-Shepherd reward model is used for verification, a large number of incorrect solutions cluster around $R_\text{correctness}=0$.
This observation corroborates our earlier hypothesis that erroneous steps typically reflect low-quality (incorrect) reasoning chains. 
Conversely, when only the PRM800K reward model is employed for verification, numerous incorrect solutions gather around $R_\text{potential}=0$, as shown in Figure~\ref{fig:three-max}(b). 
This highlights a limitation of PRM800K reward model: it struggles to select solutions with higher potential when all solution candidates are prone to errors.
As illustrated in Figure~\ref{fig:three-max}(c), once the two rewards are compounded in \MethodName, the performance in these two extreme scenarios is substantially improved.

Notably, within the incorrect solutions, the PRM~(correctness) score demonstrates a stronger predictive capability, as evidenced by the majority of points aligning along the left side of the plot. Conversely, among the correct solutions, several samples exhibit lower PRM scores.
This observation suggests that erroneous intermediate steps can still lead to correct final outcomes, particularly as contemporary models exhibit enhanced self-reflection abilities. This is likely one of the reasons why recent works, such as DeepSeek-R1~\cite{deepseekai2025deepseekr1incentivizingreasoningcapability}, have abandoned the use of the process reward model.

\subsection{Other Reward Mixing Approaches}
\begin{table*}[t]
\small
\centering
\begin{tabular}{lll}
\toprule
\textsc{Approach}                    & \textsc{MATH500} & \textsc{ProcessBench} \\
\midrule
Multiply two rewards from two models        &   0.498      &    63.0\%         \\
Train single-head model with multiplication of two rewards  &   0.514      &    64.4\%          \\
Train multi-head model then multiply &   \textbf{0.526}      &    \textbf{65.5}\%          \\
\bottomrule
\end{tabular}
\caption{Performances of different reward mixing approaches. 
% Even with extra model parameters and training data, mixing correctness and potential rewards with a neural network does not bring significant performance improvement.
}
    \label{tab:approaches}
\end{table*}

An alternative reward mixing approach involves training a model by directly using the product of the two rewards as the target label. This method simplifies the training process by consolidating the reward signals into a single objective, potentially enhancing the model's ability to learn the combined effect of both rewards. 

When using this method for training, the mixed reward label is required to be obtained by multiplying two soft labels. If the original binary labels are used, it can lead to a reward collapse phenomenon. Specifically, when one label is 0, the other label is completely overridden; whereas when one label is 1, it effectively results in only the other label being active.

As shown in Table~\ref{tab:approaches}, directly distilling the product of two models' labels achieves comparable but slightly worse results. Unlike learning from two independent labels, in this scenario, the model learns a more abstract label, which we speculate may influence the final performance.

% The multiplication of the two rewards in Eq. (5) assumes the probability independence of the two rewarding signals. In this sub-section, we study an alternative approach without this assumption.

% Considering two labels are now associated with each reasoning step, the golden answer \( y \) in Eq. (1) is a tuple containing two labels. And PRM training is conceptualizes as a classification problem that estimate the probability of a reasoning step with both correctness and potential reward labels are 1. This approach does not assume the independence of the two reward signals.

% This approach achieves comparable albeit slightly lower performance (as shown in the second line of Table 3). It demonstrates the validity of the independence assumption while also confirming the effectiveness of the two-head multi-task training.
% \subsection{Effect of Generator Model}

\subsection{The Weak to Strong Phenomenon}
We also observe a clear pattern similar to the weak-to-strong generalization phenomenon described by OpenAI \cite{burns2023weaktostronggeneralizationelicitingstrong}.

% As shown in Table~\ref{tab:approaches}, when we distill the multiplied score as a simple value prediction task on a 7B model, we achieve a 1.6\% accuracy gain on the MATH500 dataset.
 % compared to using the raw multiplied score alone
We conducted the same task as Section 6.2 but using a much larger 72B model. 
We trained a Qwen-2.5-Math-72B-Instruct model with a learning rate of \(1 \times 10^{-4}\) for 3 epoches. 
Due to computational constraints, the 72B model was fine-tuned using the LoRA approach~\cite{hu2022lora}.
The LoRA configuration was set as follows: rank \( r=16 \), scaling factor \( \text{lora\_alpha} = 32 \), dropout rate \( \text{lora\_dropout} = 0.05 \), with target modules specified as \( q\_proj, k\_proj, v\_proj \).
This model results in a best-of-N accuracy of 56.2\% in MATH500 and 67.5\% average F1 in ProcessBench, which is significant higher than the the corresponding 7B reward models.

We believe this phenomenon aligns with OpenAI's weak-to-strong generalization framework. The multiplication method can be seen as a very weak \textit{model}, and when its generated data is used to fine-tune a more capable model, we consistently observe improvements as the model’s overall reasoning capability increases.
% (which, in this case, is represented by model size).

We argue that behind the multiplication method, there must be a deeper, more fundamental signal or rule that governs reasoning. Fine-tuning a stronger model on data generated by a weak model likely helps elicit and uncover parts of this underlying capability.
% , progressively refining the model's abilities

\section{Conclusion}
In this paper, we have presented a novel framework for integrating correctness-based and potential-based reward signals to enhance stepwise supervision in mathematical reasoning tasks. By unifying feedback on both the correctness of individual steps and the likelihood of reaching the correct final answer, our approach captures complementary facets of the problem-solving process. Empirical evaluations on MATH500 and ProcessBench demonstrate that this combined reward substantially outperforms either reward type alone and achieves state-of-the-art under comparable resource constraints.

Moreover, our multi-task training strategy shows that correctness and potential rewards can be jointly modeled in a single, shared representation, further improving verification accuracy. This result confirms that the two reward signals focus on different aspects of reasoning: correctness flags stepwise errors, while potential leverages forward-looking guidance. Our findings indicate that balancing these two perspectives not only improves performance but also yields finer-grained insights into the reasoning chain.

Taken together, this work offers a scalable pipeline for constructing high-quality stepwise reward modeling data and a flexible methodology for fusing multiple reward dimensions. Future research could investigate more advanced reward mixing techniques, better integration with tree-search-based methods, and applications to broader domains requiring multi-step reasoning. We anticipate that our dual-reward paradigm will spur further innovation in fine-grained reward modeling.
% , ultimately pushing the boundaries of what LLMs can achieve in complex problem-solving settings.

\section*{Limitations}

Despite its demonstrated improvements, our framework has limitations. First, it relies on pseudo labels generated by reward models trained on Qwen-2.5-Math-7B, whose accuracy directly affects final performance. Larger or more capable models could yield more reliable labels. Second, the PRM800K dataset primarily focuses on correctness, which may not fully capture the self-reflective and adaptive reasoning of advanced LLMs. Future work might explore more flexible annotations to accommodate alternate reasoning paths. Finally, while we integrate correctness and potential signals through direct multiplication, other mixing strategies (e.g., adaptive weighting or RL-based approaches) remain underexplored and could further enhance stepwise supervision.

% \section*{Ethics Statement}

% \section*{Acknowledgements}

\bibliography{anthology,custom}
\bibliographystyle{acl_natbib}

\appendix

% \section{Example Appendix}
% \label{sec:appendix}

% This is a section in the appendix.

\end{document}